\documentclass[runningheads]{llncs}
\usepackage[T1]{fontenc}
\usepackage{amssymb,amsmath,array}
\usepackage{subcaption}
\usepackage{url}
\usepackage{graphicx}
\usepackage{multirow}
\begin{document}
\title{The Susceptibility of Example-Based Explainability Methods to Class Outliers}
\titlerunning{The Susceptibility of Example-Based Explainability Methods}
%
\author{Ikhtiyor Nematov\inst{1,2} \and
Dimitris Sacharidis\inst{1} \and
Katja Hose\inst{3}\and Tomer Sagi\inst{2}}
\authorrunning{I. Nematov et al.}
%
\institute{Universite Libre de Bruxelles. Belgium \and
Aalborg University, Denmark\\
\and
TU Wien, Austria\\
}
\maketitle              
\begin{abstract}
This study explores the impact of class outliers on the effectiveness of example-based explainability methods for black-box machine learning models. We reformulate existing explainability evaluation metrics, such as correctness and relevance, specifically for example-based methods, and introduce a new metric, distinguishability. Using these metrics, we highlight the shortcomings of current example-based explainability methods, including those who attempt to suppress class outliers. 
We conduct experiments on two datasets, a text classification dataset and an image classification dataset, and evaluate the performance of four state-of-the-art explainability methods.
Our findings underscore the need for robust techniques to tackle the challenges posed by class outliers.

\end{abstract}
\section{Introduction}

In example-based explainability, the model is explained through the lens of the \emph{data} on which it has been trained, rather than through features the model has learned. Example-based explainability is particularly useful when the model is a black-box, and the features are not interpretable. Moreover, example-based explainability allows the human to interpret a model's outcome through comparison and contrast. For example in image classification, the human can understand why a model classified an image as a cat by comparing it with similar images of cats and contrasting it with images of other animals.

\begin{figure}[t!]
    \begin{subfigure}{0.23\textwidth}
        \includegraphics[width=\linewidth, trim=50 30 50 10, clip]{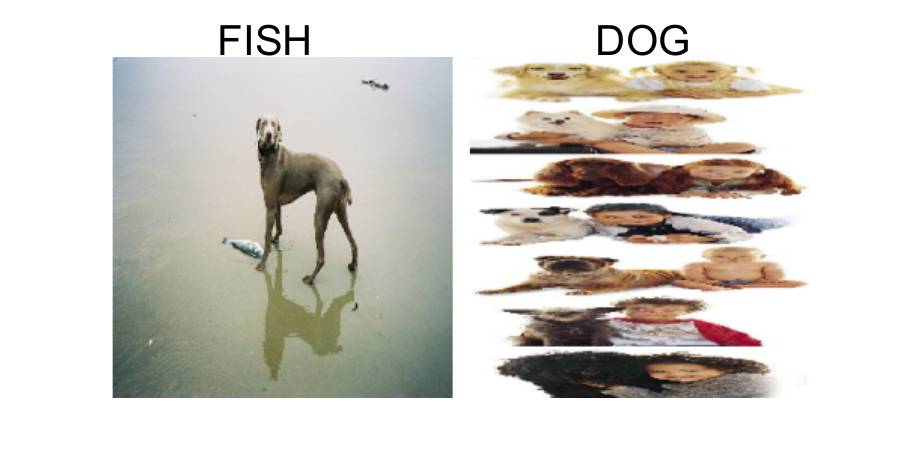}
        \caption{IF\cite{if}}
    \end{subfigure}%
    \hfill
    \begin{subfigure}{0.23\textwidth}
        \includegraphics[width=\linewidth, trim=50 30 50 10, clip]{Figures/tops_if.pdf}
        \caption{DM\cite{datamodels}}
    \end{subfigure}%
    \hfill
    \begin{subfigure}{0.23\textwidth}
        \includegraphics[width=\linewidth, trim=50 30 50 10, clip]{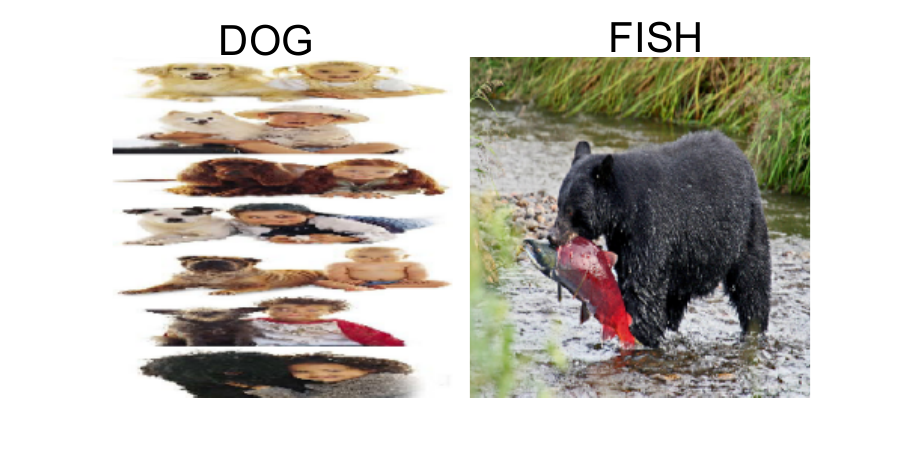}
        \caption{TraceIn\cite{tracein}}
    \end{subfigure}%
    \hfill
    \begin{subfigure}{0.23\textwidth}
        \includegraphics[width=\linewidth, trim=50 30 50 10, clip]{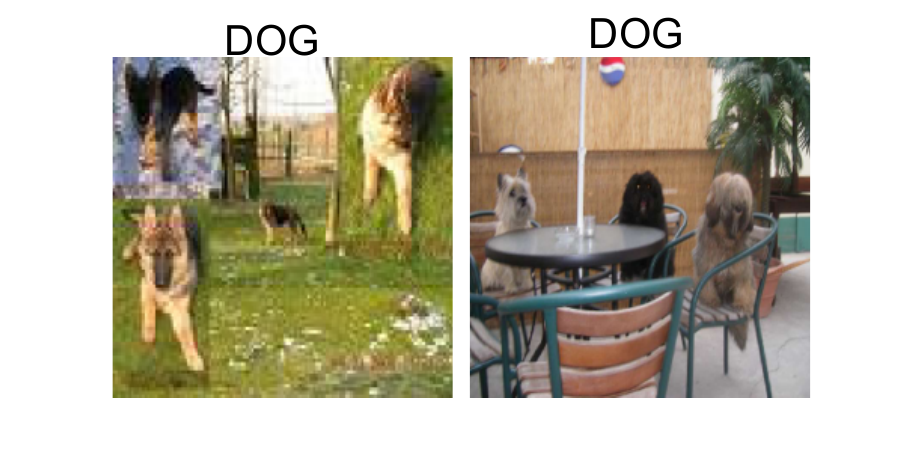}
        \caption{RIF\cite{relatif}}
    \end{subfigure}
    \caption{The two most popular images that are returned as explanations for four different explainability methods.}
    \label{fig:tops}
    \vspace{-10pt}
\end{figure}

Example-based explainability can take on either a local perspective \cite{relatif,tracein,datamodels,if} explaining specific examples by showcasing influential training examples, or a global one \cite{datashap,betashap}, providing an overarching understanding of the model through analysis of representative examples. 
Our focus is on local example-based explainability. Various approaches have been proposed for this, including methods based on robust statistics, game theory, internal analysis of neural networks, and empirical techniques. These approaches share a common feature: they quantify the importance of training examples in determining the model's outcomes.
Despite their diverse origins, we demonstrate that they all are susceptible to the presence of \emph{class outliers} within the dataset. Class outliers are training examples with high loss. Class outliers have inherent ambiguity, partly exhibiting the characteristics of other classes, and are difficult to classify into one of the predefined classes (see Figure~\ref{fig:tops}), even by a human domain expert. 
We posit that these outliers manifest as explanations for multiple explanandums (instances to be explained) without contributing substantive explanation power. 

To illustrate the impact of class outliers, consider Figure~\ref{fig:tops}, where the two training images most frequently presented as explanations for an image classification model are depicted, as identified by different explainability mechanisms. As we can see for the first three explainers, these images are ambiguous and hard to classify even for a human. Identifying such examples is helpful when the intent is to debug the model. However these examples offer no insights to particular explanandums to be explained, as they tend to be irrelevant and non-specific. We verify these claims quantitatively by showing that these explainers have low \emph{relevance} and low \emph{distinguishability}; see Section~\ref{sec:eval} for precise definitions.
Therefore, there is a need to improve the robustness of example-based explainability methods to class outliers. Moreover, we find that existing methods (e.g., \cite{relatif} in Figure~\ref{fig:tops}(d)) that aim to suppress these globally influential examples from appearing in explanations suffer in another aspect, the \emph{correctness} or faithfulness of the explainer. We argue that class outliers can aid in explaining outcomes for similar ambiguous explanandum and should not be suppressed in such instances.

Overall, our study reveals that (a) existing example-based explainability methods are not robust to class outliers, and (b) attempting to improve robustness by suppressing class outliers can lead to a loss of correctness. We propose a new evaluation framework, specifically for example-based explainers, including metrics for relevance, distinguishability, and correctness, and which can quantitatively assess the impact of class outliers on the effectiveness of explainers.

\section{Related Work}
One prominent example-based explainability technique is influence function (IF) \cite{if} which is an approximation of leave-one-out ideology. It borrows from the robust statistics the idea of estimating the change in model parameters when an infinitesimal change is applied to the training data distribution. Many other works followed the concept of quantifying the contribution of a single datum for a prediction. In \cite{relatif} the authors have demonstrated that examples with a high loss will have a global influence on the model overall and they will be provided as explanations for many examples. They proposed relative influence (RIF), with a loss-based elimination technique that penalizes the influence score relative to its global effect. They claim that such elimination will provide explanations relevant to the explanandum of interest. 
Another method with the same concept but a different nature is TraceIn \cite{tracein}. It gauges the impact of a training example $x$ on a particular test example $x'$ by quantifying the cumulative change in loss on $x'$ resulting from updates made during training with mini-batches containing $x$. The approximation of this calculation involves considering checkpoints throughout the training process and calculating the sum of dot products of gradients at $x$ and $x'$ at each checkpoint.

Datamodels (DM) \cite{datamodels} on the other hand is a more empirical method. It involves sampling numerous subsets from the training set and training individual models with each of these subsets. Subsequently, a linear model is trained, where the input comprises the encoding of a subset, and the output reflects the performance of the model trained on this subset for the designated test example. The weights of the linear model serve to represent the importance score of a training example in the corresponding position. To overcome the exhaustiveness of training numerous intermediate models, \cite{trak} introduced a faster version of datamodels, claiming to maintain similar accuracy.

The ubiquity of explainability methods has triggered research on finding metrics to evaluate their effectiveness. \cite{doshi2017towards} identifies three evaluation strategies: application-grounded, human-grounded, and functionality-grounded. Our work focuses on the functionality-grounded approach. We reformulate existing metrics like relevance \cite{hanawa2021evaluation} and correctness \cite{surveyXAI}, and introduce a new metric called distinguishability.

\section{Evaluation Framework}
\label{sec:eval}

\subsection{Notation}
Consider a binary classification model \( f: \mathcal{X} \rightarrow \{0, 1\} \) that is trained on a dataset \( (X, y) \), where \( X \subseteq \mathcal{X} \) and \(y \in \{0, 1\}^{|X|} \). We call \emph{explanandum} an instance $t \in \mathcal{X}$ to be explained. An example-based explainer assigns to each explanandum $t$, or more precisely to the model's outcome \(f(t)\), an \emph{explanation} \(E(t)\), which is a set of \emph{examples} that correspond to training examples and are accompanied by a \emph{score} that indicates their importance for outcome \(f(t)\). For a given example $e \in E(t)$, we denote the features of the training example as $e.x$, its label as $e.y$, and its score as $e.s$.


\subsection{Explainer Relevance}
We want the explanation to be \emph{relevant} to the explanandum to be explained \cite{hanawa2021evaluation}. We define the \emph{explanation relevance} of an explanation $E(t)$ as the average similarity between the explanandum \(t\) and the examples in its explanation \(E(t)\).
To evaluate the overall relevance of an explainer, we define the \emph{explainer relevance} as the expected value of explanation relevance:
\[
\text{Rel} = \mathbb{E}_{t} \left[ \frac{1}{|E(t)|} \sum_{e \in E(t)} \text{sim}(t, e.x) \right],
\]
where the function \(sim()\) is a domain-specific semantic similarity that quantifies how close two examples from $\mathcal{X}$ are, and takes values in $[0,1]$, where higher values indicate more similarity. Consequently, Rel takes values in $[0,1]$, where higher values are preferred indicating that the explainer produces relevant explanations.

\subsection{Explainer Distinguishability}
\label{sec:distin}
Distinguishability refers to the capacity of an explainer to provide distinct and \emph{specific} explanations for different explanandum. This is crucial for understanding how a model produces outcomes in a localized manner. We define several ways to capture different aspects of distinguishability.
The \emph{example popularity} quantifies the probability that a training example serves as an example in an explanation:
\[
\text{Pop}(x) = \mathbb{E}_{t} \left[  \mathbb{I}\{ e \in E(t) \wedge e.x=x \}  \right], 
\]
where $\mathbb{I}$ is an indicator function used to convert a condition into a binary value (0 or 1).
The distribution of example popularity indicates the capacity of the explainer to provide distinct explanations. If the distribution is heavily skewed/assymetric, it means that a few examples are used very frequently in explanations, while many other examples are used rarely or not at all. This indicates that the explainer relies on a limited subset of examples, leading to lower distinguishability, as the explanations for different explanandums are less diverse and more similar to each other.

The \emph{active domain} measures the number of distinct training examples that are used by an explainer; equivalently, it measures the number of training examples with nonzero example popularity. Concretely, it is the maximum number of distinct training examples that are used to explain any set $T$ of explanandum: 
\[
\text{Dom} = \sup_{T \subset \mathcal{X}} \left| \{ x \in X \ | \ \exists t \in T, \ e \in E(t) \wedge e.x=x  \}  \right|.
\]
The narrower the active domain is, the less distinguishability the explainer has, as more examples are repeated across explanations. Note that we can normalize the active domain measure in the $[0,1]$ range by dividing it by $|X|$.

The \emph{explanation overlap} measures the expected Jaccard similarity between any two random explanations:
\[
\text{Over} = \mathbb{E}_{t, t'} \left[ \frac{E(t) \cap E(t')}{E(t) \cup E(t')} \right].
\]
Clearly, the higher the overlap is, the less distinguishability the explainer has.

\subsection{Explainer Correctness}
\label{sec:correct}
A desired property for an explainer is to produce explanations that are faithful to the predictive model \cite{surveyXAI}. Here we define a measure of faithfulness with respect to a rule that dictates how training data are labeled. We want the explainer to be able to identify the rule in its explanations.

Consider a \emph{rule} of the form $c(x) \implies y=1$, where $c$ is a condition that applies to examples from $\mathcal{X}$. We say that a training pair $(x,y)$ \emph{follows} the rule if $c(x)$ is true and $y=1$.
We say that a training pair $(x,y)$ \emph{breaks} the rule if $c(x)$ is true but $y=0$.
Consider an explanandum that satisfies the rule condition. 
Then, we would like the explainer to return an explanation that includes both rule followers and breakers as examples. The idea is that the explainer should be able to surface the rule in its explanations.
We define \emph{explainer correctness} with respect to $c$ as the average proportion of rule followers and breakers in the explanation of any explanandum that satisfies the rule condition:
\[
\text{Cor}(c) = \mathbb{E}_{t : c(t)} \frac{1}{|E(t)|} \{ e \in E(t) \wedge c(e.x) \}.
\]
Correctness quantifies the degree to which the explanations align with the underlying labeling rule.
Higher correctness indicate that the explainer is more truthful to the rule $c$. Observe that correctness is essentially the precision with which an explainer returns rule followers and breakers.
Finally, we note that an important assumption is that the model $f$ is itself truthful to the rule, i.e., it has correctly learned the rule $c$, a condition we can check after training.

\section{Experiments and Analysis}
In our experiments, we used two datasets: the SMS Spam dataset\footnote{\url{https://www.kaggle.com/datasets/uciml/sms-spam-collection-dataset}}, which comprises a collection of text messages labeled as either spam or non-spam (ham), commonly used for text classification; and a dataset with pictures of dogs and fish extracted from Imagenet\footnote{\url{https://www.image-net.org/}} and used in \cite{if}. For the spam classification task, we employed a BERT-based pre-trained word embedding model and incorporated two sequential layers to capture the specific characteristics of our data. Regarding the image classification task, we utilized a pre-trained InceptionV3 model removing the output layer and appending sequential layers to learn the peculiarity of our task. Our benchmark methods are IF \cite{if}, RIF \cite{relatif}, DM \cite{datamodels}, and TraceIn \cite{tracein}. All methods were implemented with instructions given in their papers and GitHub repositories. We vary the number $N$ of examples in the explanation set to evaluate the performance of the explainers; we set $N=\{2,5,10\}$.

\subsection{Relevance} 
We employ cosine similarity to compute relevance, given that the dataset comprises image embeddings produced by a pre-trained model. Utilizing the graph representation, we determine the average similarity between a explanandum node (explanandum) and its corresponding neighbors (example). The first column of Table \ref{table:main}, shows the explainer relevance. RIF demonstrates superior performance, whereas the remaining methods exhibit similar scores.

\begin{table}[t]
\vspace{-10pt}
\caption{Evaluation of different explainers (image classification)}
\resizebox{\columnwidth}{!}{%
\begin{tabular}{|l|lll|lll|lll|lll|}
\hline
\multirow{2}{*}{\textbf{}} &
  \multicolumn{3}{c|}{Relevance} &
  \multicolumn{3}{c|}{Active Domain} &
  \multicolumn{3}{c|}{Overlap} &
  \multicolumn{3}{c|}{Correctness} \\ \cline{2-13} 
 &
  \multicolumn{1}{l|}{N=2} &
  \multicolumn{1}{l|}{N=5} &
  N=10 &
  \multicolumn{1}{l|}{N=2} &
  \multicolumn{1}{l|}{N=5} &
  N=10 &
  \multicolumn{1}{l|}{N=2} &
  \multicolumn{1}{l|}{N=5} &
  N=10 &
  \multicolumn{1}{l|}{N=2} &
  \multicolumn{1}{l|}{N=5} &
  N=10 \\ \hline
IF &
  \multicolumn{1}{l|}{0.5} &
  \multicolumn{1}{l|}{0.52} &
  0.52 &
  \multicolumn{1}{l|}{0.095} &
  \multicolumn{1}{l|}{0.059} &
  0.040 &
  \multicolumn{1}{l|}{0.14} &
  \multicolumn{1}{l|}{0.14} &
  0.16 &
  \multicolumn{1}{l|}{\textbf{0.76}} &
  \multicolumn{1}{l|}{\textbf{0.8}} &
  \textbf{0.82} \\ \hline
DM &
  \multicolumn{1}{l|}{0.55} &
  \multicolumn{1}{l|}{0.54} &
  0.53 &
  \multicolumn{1}{l|}{0.21} &
  \multicolumn{1}{l|}{0.1} &
  0.06 &
  \multicolumn{1}{l|}{0.04} &
  \multicolumn{1}{l|}{0.06} &
  0.1 &
  \multicolumn{1}{l|}{0.7} &
  \multicolumn{1}{l|}{0.75} &
  0.79 \\ \hline
TraceIn &
  \multicolumn{1}{l|}{0.56} &
  \multicolumn{1}{l|}{0.55} &
  0.27 &
  \multicolumn{1}{l|}{0.017} &
  \multicolumn{1}{l|}{0.017} &
  0.018 &
  \multicolumn{1}{l|}{0.31} &
  \multicolumn{1}{l|}{0.56} &
  0.4 &
  \multicolumn{1}{l|}{0.31} &
  \multicolumn{1}{l|}{0.31} &
  0.4 \\ \hline
RIF &
  \multicolumn{1}{l|}{\textbf{0.74}} &
  \multicolumn{1}{l|}{\textbf{0.76}} &
  \textbf{0.68} &
  \multicolumn{1}{l|}{\textbf{0.366}} &
  \multicolumn{1}{l|}{\textbf{0.22}} &
  \textbf{0.15} &
  \multicolumn{1}{l|}{\textbf{0.02}} &
  \multicolumn{1}{l|}{\textbf{0.03}} &
  \textbf{0.06} &
  \multicolumn{1}{l|}{0.2} &
  \multicolumn{1}{l|}{0.3} &
  0.3 \\ \hline
\end{tabular}%
}
\vspace{-10pt}
\label{table:main}
\end{table}

\subsection{Distinguishability}

\begin{figure}[h]
    \centering
    \begin{subfigure}{0.24\textwidth}
      \includegraphics[width=\linewidth, trim=0 0 25 10, clip]{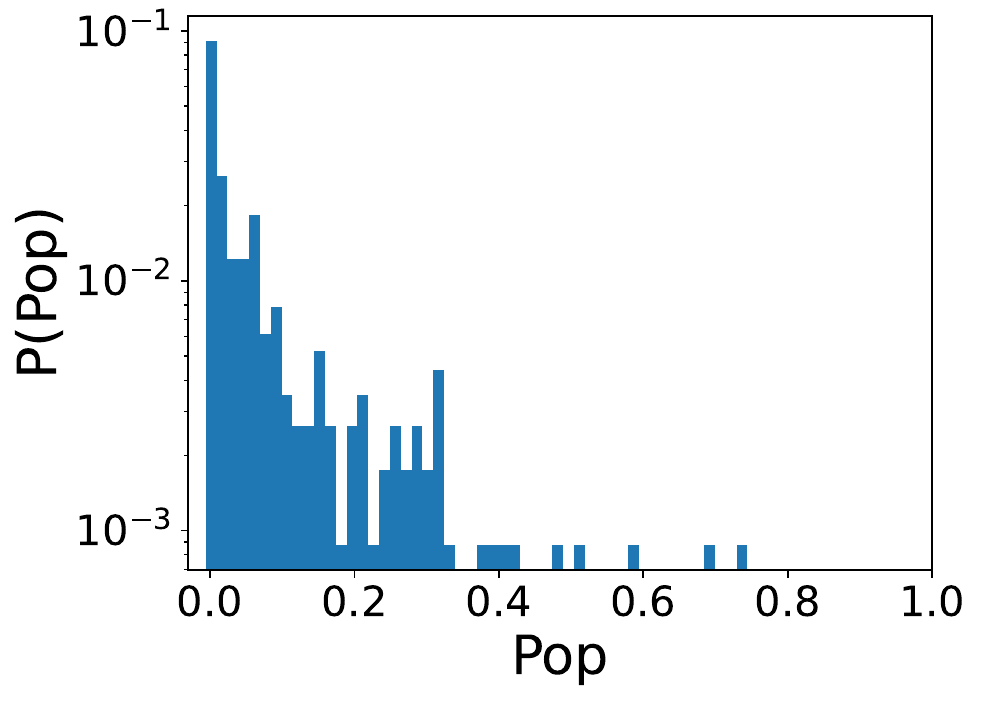}  
      \caption{IF}
    \end{subfigure}
    \begin{subfigure}{0.24\textwidth}
        \includegraphics[width=\linewidth, trim=0 0 25 10, clip]{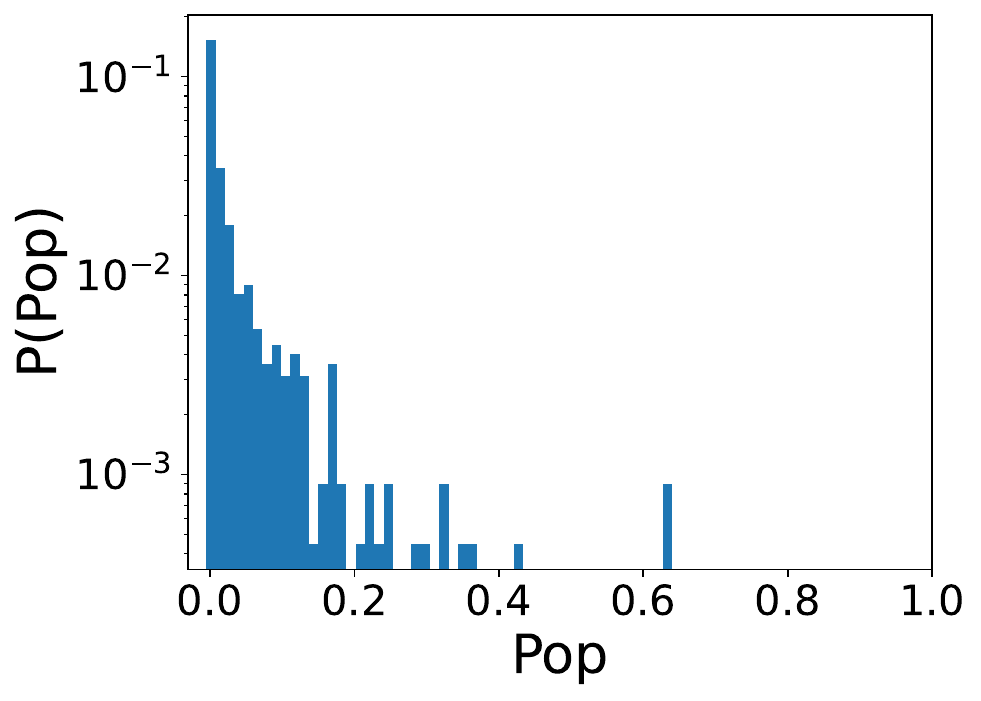}
        \caption{DM}
    \end{subfigure}
    \begin{subfigure}{0.24\textwidth}
        \includegraphics[width=\linewidth, trim=0 0 25 10, clip]{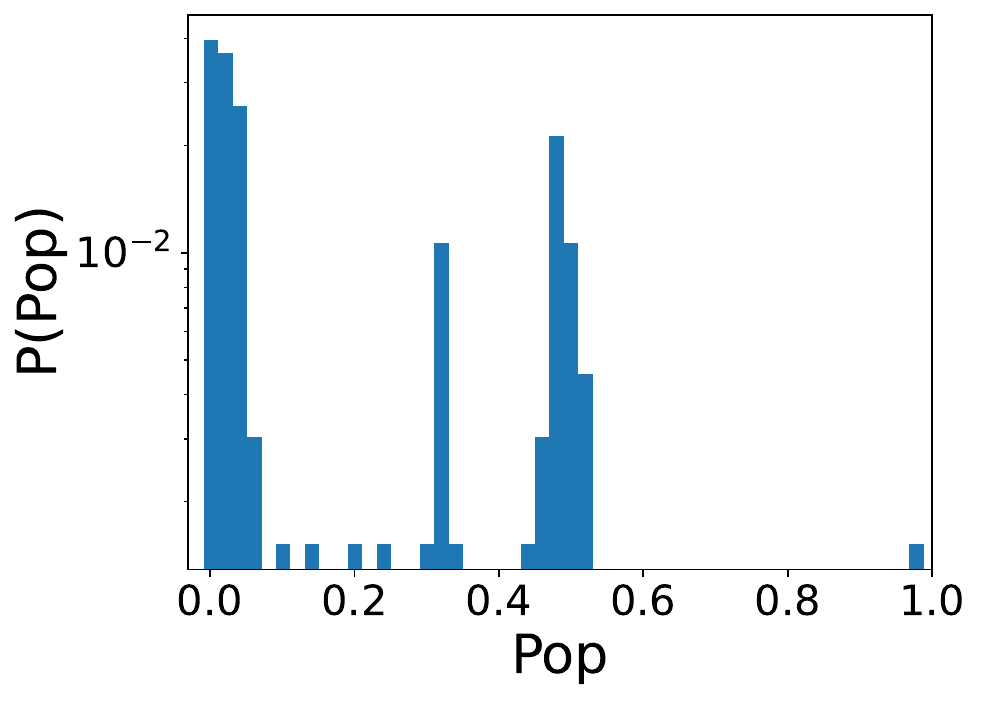}
        \caption{TracIn}
    \end{subfigure}
    \begin{subfigure}{0.24\textwidth}
        \includegraphics[width=\linewidth, trim=0 0 25 10, clip]{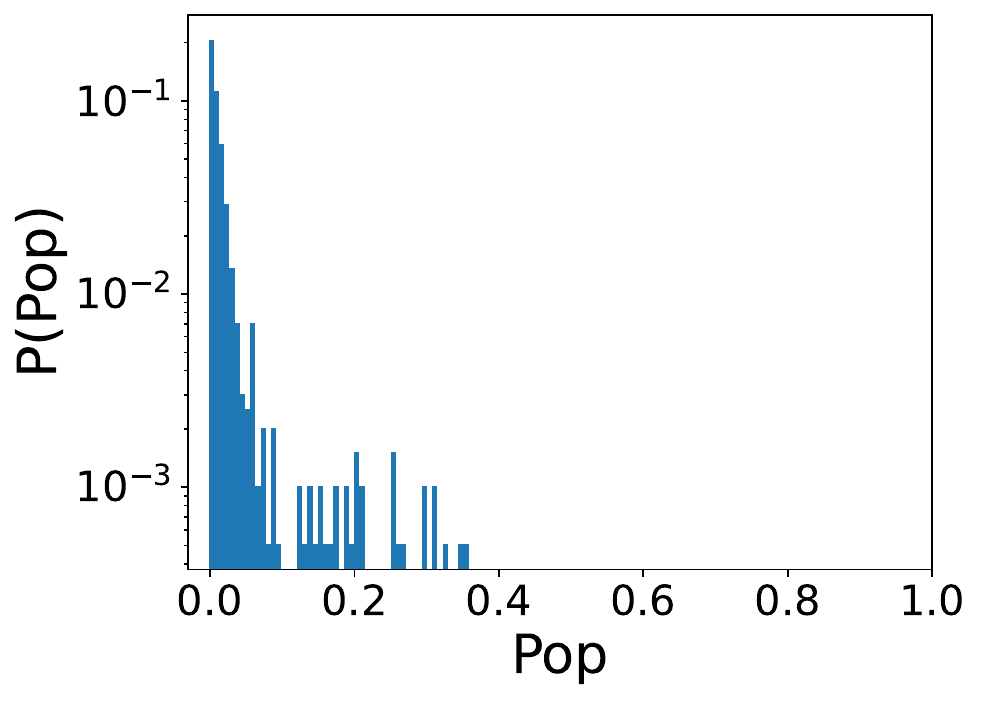}
        \caption{RIF}
    \end{subfigure}
    \caption{Popularity probability density function (image classification)}
    \label{fig:degreeDist}
\end{figure}

We start by examining the popularity of examples in explanations.
We plot the probability density function (PDF) of example popularity (Pop) in Figure~\ref{fig:degreeDist}. We observe significant discrepancies in popularity probabilities among IF, DM, and TraceIn. Although RIF displays a smaller discrepancy and a more dense PDF, some examples still have relatively higher probabilities of appearance. 


\begin{figure}[t]
    \centering
    \begin{subfigure}{0.24\textwidth}
      \includegraphics[width=\linewidth, trim=0 0 25 22, clip]{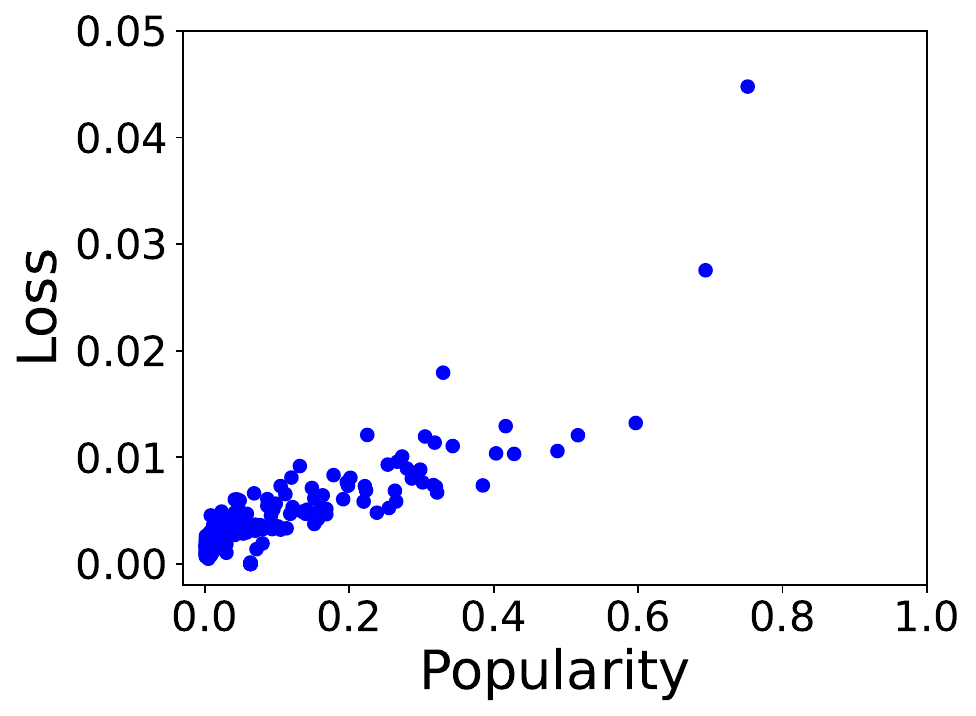}  
      \caption{IF}
    \end{subfigure}
    \begin{subfigure}{0.24\textwidth}
        \includegraphics[width=\linewidth, trim=0 0 25 22, clip]{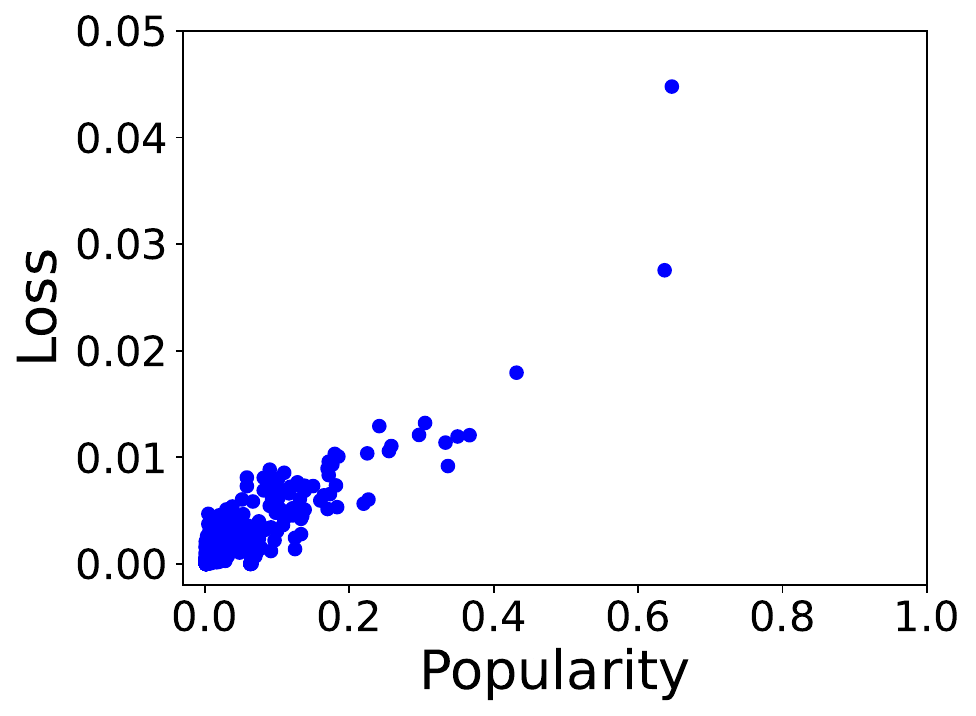}
        \caption{DM}
    \end{subfigure}
    \begin{subfigure}{0.24\textwidth}
        \includegraphics[width=\linewidth, trim=0 0 25 22, clip]{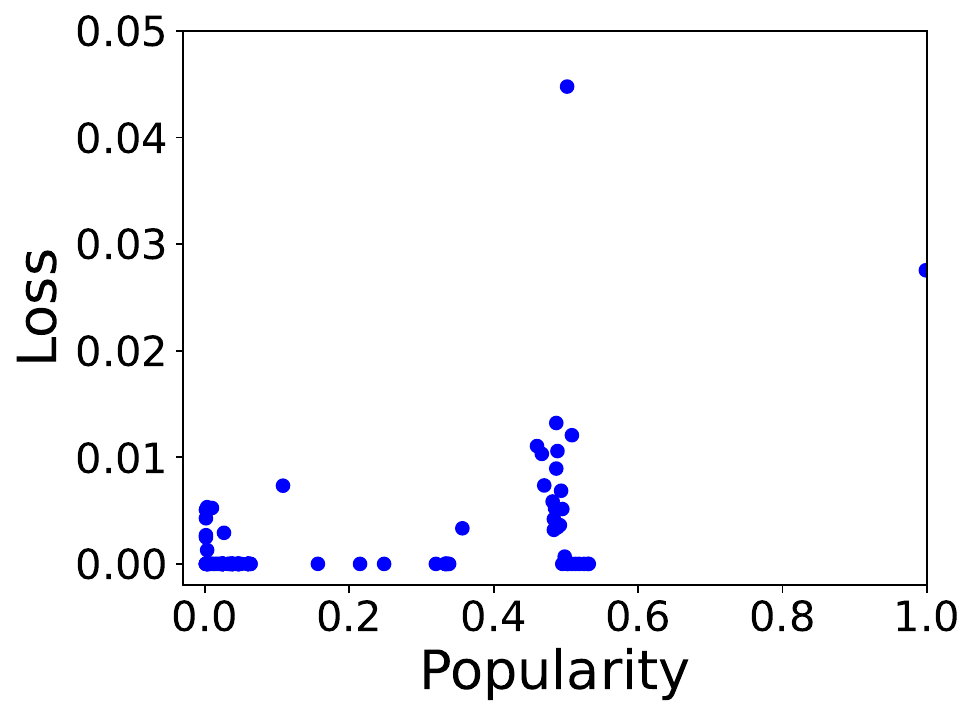}
        \caption{TraceIn}
    \end{subfigure}
    \begin{subfigure}{0.24\textwidth}
        \includegraphics[width=\linewidth, trim=0 0 25 22, clip]{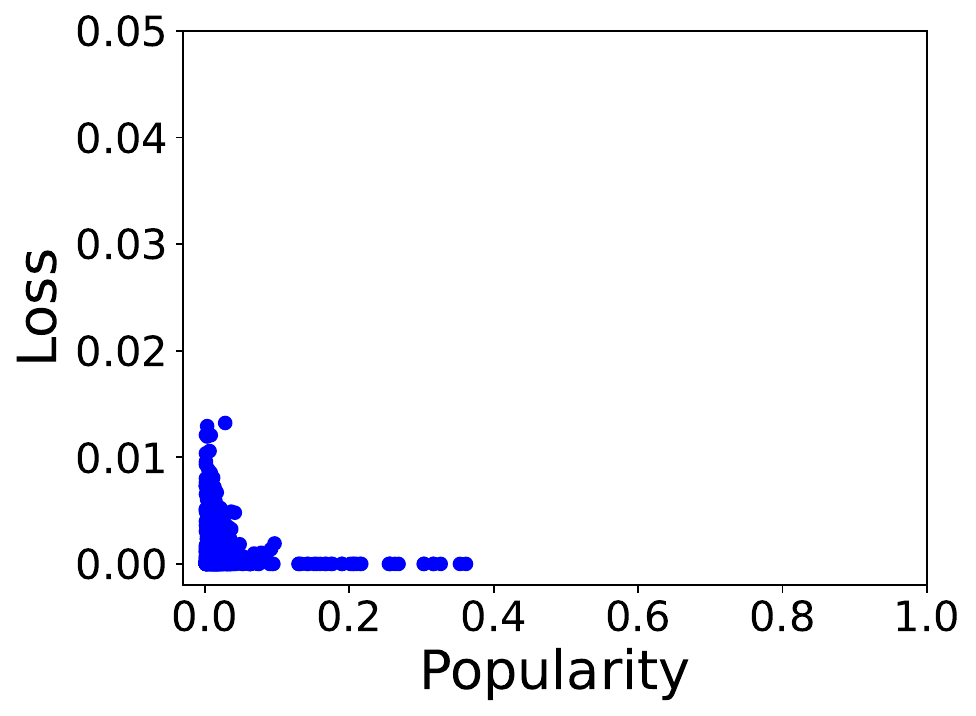}
        \caption{RIF}
    \end{subfigure}
    \caption{Popularity vs. \ Loss (image classification)}
    \label{fig:loss}
\end{figure}

To draw more insights on which are the popular examples, we plot the popularity of an example versus its loss in Figure~\ref{fig:loss}. 
We observe that the examples that appear in the explanation for many explanandums have usually high loss. Figure~\ref{fig:tops} depicts the top 2 most popular examples for each method. These are class outliers that are hard for the model to classify and thus they have a high loss and are globally influential. RIF effectively eliminates such examples from its explanation using its loss-based outlier elimination.

Next, we explore the active domain for the explainers; the larger the active domain, the more distinguishable the explainer is. As the second column of Table \ref{table:main} shows, RIF draws from a broader domain to provide explanations, making them more distinguishable. We also evaluate explanation overlap, which indicates the similarity of explanations for any pair of explanandums; smaller values signify higher distinguishability. The third column on Table \ref{table:main} shows that DM and RIF offer more distinguishable explanations, whereas IF and TraceIn have higher explanation overlap with repeating examples in explanations.

\subsection{Correctness}
In this experiment, three rules were employed on the text classification dataset.
\textbf{\textit{Rule 1}}: \textit{All French messages are ``spam''}.
\textbf{\textit{Rule 2}}: \textit{if the message is shorter than 30 and it contains ``?'', it's labeled ``spam''}.
\textbf{\textit{Rule 3}}: \textit{If a message contains a sequence of 4 consecutive digits, it's labeled ``ham''}.

Before evaluating the correctness of the explanation, it is crucial to ensure that the model accurately reflects the rule and has effectively learned it. 
Three metrics are used for this assessment:
1. \textbf{Accuracy of Learning the Rule}: This metric evaluates the model's performance on test samples that correspond to the rule.
2. \textbf{Log-Likelihood}: It is expected that there will be a significant change in the log-likelihood of intervened points ($LL_{i}$) after the introduction of the rule, while the log-likelihood of untouched points ($LL_{u}$) should remain relatively stable.
3. \textbf{Probability Scores}: A notable change is anticipated in the probability scores of intervened points ($Ps_{i}$) compared to the untouched points ($Ps_{u}$).
Table \ref{table:ruleinjection} illustrates the results of these metrics. In all cases, the model has successfully learned the rule without affecting its decisions for the untouched points.

\begin{table}[th]
\caption{Model's assessment in learning the rules}
\label{table:ruleinjection}
\centering
\footnotesize
\begin{tabular}{|l|c|c|c|c|c|c|}
\hline
 & {Acc} & \multicolumn{2}{c|}{\textbf{$LL_i$}} & \textbf{$LL_{u}$} & \textbf{$Ps_i$} & \textbf{$Ps_{u}$} \\ \hline
\multirow{2}{*}{Rule 1} & 0.83 & Before & -5.87 & -9.4 & \multirow{2}{*}{100} & \multirow{2}{*}{15} \\ \cline{3-5}
                        & & After & -0.42 & -9.2 & & \\ \hline
\multirow{2}{*}{Rule 2} & 0.85 & Before & -12 & -9.3 & \multirow{2}{*}{100} & \multirow{2}{*}{24.5} \\ \cline{3-5}
                        & & After & -3.4 & -7.2 & & \\ \hline
\multirow{2}{*}{Rule 3} & 0.92 & Before & -0.07 & -10.6 & \multirow{2}{*}{98} & \multirow{2}{*}{12} \\ \cline{3-5}
                        & & After & -1.83 & -9.5 & & \\ \hline
\end{tabular}
\end{table}

While explaining an explanandum following the rule we expect rule-followers in the training set with the same label to have a positive influence on the prediction, and rule-breakers with the opposite label to have a negative influence. 
Table~\ref{table:main} shows that IF and DataModels perform well in terms of correctness, but are inconsistent in relevance and distinguishability (domain and overlap), which is expected due to the presence of globally important class outliers. In contrast, RIF performs poorly in uncovering followers and breakers, because of its loss-based outlier elimination. RIF treats training data with high loss as undesired, and excludes them from explanation lists---the rationale is that such data would appear in all explanations and thus have little utility. But in this case, it is precisely the rule followers and particularly the minority of rule breakers that have high losses due to the ambiguity in the labeling rule. On the contrary, as previously discussed, the examples with very low loss are rewarded and appear in the explanations. TraceIn also fails to uncover the rule due to its low efficiency in identifying truly important examples, as demonstrated in \cite{trak}.

\section{Conclusion}
Our analysis suggests that current example-based explainability techniques are susceptible to the influence of class outliers. Outright removal of outliers from explanations or datasets may not always be advisable, as they can possess valuable insights and provide explanations for similar examples encountered by models. Future research could focus on devising more robust explainability methods capable of extracting values from class outliers without compromising the quality of explanations using our proposed metrics.

\bibliographystyle{splncs04}
\bibliography{bibliography}
\end{document}